\documentclass[12pt]{article}

\usepackage{rsipacks} % use rsipacks.sty to add more packages

\usepackage{float}  
\usepackage{subcaption}

\usepackage[font=it]{caption}
\theoremstyle{definition}
\usepackage{graphicx}
\theoremstyle{remark}
\sectionfont{\large \centering}
\subsectionfont{\normalsize \centering}
\crefname{claim}{claim}{claims}
\Crefname{claim}{Claim}{Claims}
\crefname{app-corollary}{corollary}{corollaries}
\Crefname{app-corollary}{Corollary}{Corollaries}
\crefname{app-definition}{definition}{definitions}
\Crefname{app-definition}{Definition}{Definitions}
\crefname{figure}{figure}{figures}
\Crefname{figure}{Figure}{Figures}
\crefname{lemma}{lemma}{lemmata}
\Crefname{lemma}{Lemma}{Lemmata}
\crefname{app-lemma}{lemma}{lemmata}
\Crefname{app-lemma}{Lemma}{Lemmata}
\crefname{app-proposition}{proposition}{proposition}
\Crefname{app-proposition}{Proposition}{Proposition}
\crefname{app-theorem}{theorem}{theorems}
\Crefname{app-theorem}{Theorem}{Theorems}
\begin{document}
\title{LungX: A Hybrid EfficientNet–Vision Transformer Architecture with Multi-Scale Attention for Accurate Pneumonia Detection}
\author{Mansur Yerzhanuly}
\date{}
\maketitle

\begin{abstract}
\noindent Pneumonia remains a leading global cause of mortality where timely diagnosis is critical. We introduce LungX, a novel hybrid architecture combining EfficientNet's multi-scale features, CBAM attention mechanisms, and Vision Transformer's global context modeling for enhanced pneumonia detection. Evaluated on 20,000 curated chest X-rays from RSNA and CheXpert, LungX achieves state-of-the-art performance ($86.5\%$ accuracy, 0.943 AUC), representing a $6.7\%$ AUC improvement over EfficientNet-B0 baselines. Visual analysis demonstrates superior lesion localization through interpretable attention maps. Future directions include multi-center validation and architectural optimizations targeting  $> 88\%$ accuracy for clinical deployment as an AI diagnostic aid.

\end{abstract}

% This is paper.tex
\section{Introduction}
\noindent Pneumonia is a major cause of morbidity and mortality worldwide, particularly affecting young children, the elderly, and immunocompromised patients. According to the World Health Organization, pneumonia accounts for millions of hospitalizations annually and remains a leading infectious cause of death in children under five years of age \cite{who2020pneumonia}. Chest X-ray (CXR) imaging is the most common diagnostic tool for pneumonia due to its cost-effectiveness and wide availability. However, accurate interpretation of CXR images requires significant expertise and is subject to inter-observer variability, especially in cases with subtle or diffuse lung opacities \cite{rajpurkar2017chexnet}. As illustrated in Figure 1, pneumonia may present with either localized bacterial consolidations or diffuse viral opacities, making visual diagnosis challenging even for experienced radiologists. These challenges are further amplified in resource-limited settings where trained radiologists may be scarce.
\begin{figure}[H]
\centering
\includegraphics[width=0.9\linewidth]{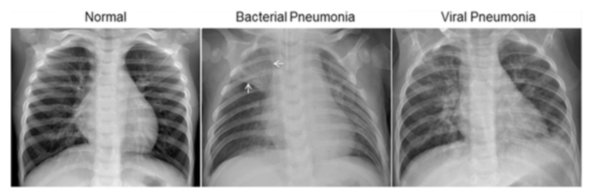}
\caption{Representative chest X-rays showing (a) normal lungs, (b) bacterial pneumonia with localized consolidation, and (c) viral pneumonia with diffuse interstitial opacities. These examples highlight the variability of pneumonia manifestations that make accurate visual diagnosis challenging. \cite{kermany2018identifying}}
\label{fig:pneumonia_examples}
\end{figure}
\noindent
Deep learning has emerged as a promising approach for automated pneumonia detection, with convolutional neural networks (CNNs) achieving radiologist-level performance in large-scale studies \cite{irvin2019chexpert}. While CNNs excel at extracting local image features, they often struggle to integrate the broader structural context of the lungs. This limitation is critical for pneumonia detection, as pathological signs may range from small, localized opacities to widespread consolidation patterns spanning large lung regions. Attention mechanisms and hybrid CNN–Transformer architectures have been proposed to address these issues, but many still lack efficient multi-scale feature fusion and robustness to variations in dataset distribution and class imbalance \cite{tan2019efficientnet}\cite{woo2018cbam}.\\\\
To address these challenges, we propose EffViT-AttnNet, a hybrid architecture combining the parameter-efficient EfficientNet backbone with Convolutional Block Attention Modules (CBAM) for spatial and channel-wise refinement at multiple scales \cite{7404017}. Mid-level feature maps from multiple stages are fused using a multi-scale fusion block, which is then processed by a pretrained DeiT Vision Transformer to capture long-range dependencies across the lung fields \cite{touvron2021training}. To enhance generalization, we employ aggressive data augmentation, class-balanced sampling, and a combined Binary Cross-Entropy + Focal loss to mitigate class imbalance \cite{lin2017focal}.\\\\
Our approach is evaluated on a combined dataset from the RSNA Pneumonia Challenge and CheXpert, leveraging both diverse acquisition sources and heterogeneous patient populations \cite{rajpurkar2018chexnext}. Experimental results demonstrate that EffViT-AttnNet achieves high accuracy and robustness while maintaining computational efficiency, making it suitable for deployment in varied clinical environments.\\\\
The main contributions of this work are:

A multi-scale hybrid CNN–Transformer model for pneumonia detection, integrating EfficientNet backbones, CBAM attention, and DeiT Transformers.

A feature fusion mechanism that unifies mid-level multi-scale features before global context modeling.

A training strategy incorporating strong augmentation, balanced sampling, and combined loss functions to enhance robustness on imbalanced, heterogeneous datasets.

Comprehensive evaluation on merged RSNA and CheXpert datasets, demonstrating superior performance compared to conventional CNN or Transformer-only approaches.\\\\

\section{Related Works} \label{sec:setting}

\noindent Pneumonia remains a leading cause of global morbidity and mortality, especially among vulnerable populations such as children and the elderly. Early and accurate detection is critical to improving treatment outcomes. Chest X-ray (CXR) imaging is the most widely used diagnostic tool; however, interpretation relies heavily on radiologist expertise, which is time-intensive, prone to inter-observer variability, and often unavailable in low-resource settings \cite{irvin2019chexpert}. Recent advances in deep learning have enabled automated CXR analysis, with several models achieving radiologist-level performance in pneumonia detection.\\\\
\textbf{2.1. CNN-Based Pneumonia Detection}\\\\
The availability of large-scale datasets such as ChestX-ray8 \cite{8099852} and CheXpert \cite{irvin2019chexpert} has catalyzed the development of convolutional neural networks (CNNs) for thoracic disease classification. CheXNet \cite{rajpurkar2017chexnet}, a DenseNet-121 model trained on NIH CXR data, was among the first to match expert-level accuracy in pneumonia detection. Transfer learning approaches have since reduced training time while improving performance on domain-specific datasets such as the RSNA Pneumonia Challenge \cite{rsna2018pneumonia} \cite{hashmi2020efficient}. Despite their success, pure CNN models often struggle with small or diffuse lesions due to their inherently local receptive fields.\\\\
\textbf{2.2. Attention Mechanisms in Medical Imaging}\\\\
Attention modules have been proposed to improve lesion localization by guiding models to focus on diagnostically relevant regions. The Convolutional Block Attention Module (CBAM) \cite{woo2018cbam} enhances both spatial and channel attention, while \cite{guan2018diagnose} designed an attention-guided CNN to mimic radiologist decision-making. More recent works such as PCXRNet \cite{9705095} combine multi-convolution blocks with condensed attention to achieve higher diagnostic accuracy. However, most of these designs operate at a single feature scale, potentially overlooking multi-scale pneumonia patterns ranging from small focal opacities to large lobar consolidations.\\\\
\textbf{2.3. Hybrid CNN–Transformer Architectures}\\\\
While CNNs capture local textures effectively, Transformers excel at modeling long-range dependencies. Vision Transformers (ViTs) \cite{dosovitskiy2021an} and variants such as DeiT \cite{touvron2021training} have shown promise in medical imaging by capturing global structural patterns across the entire lung field. Hybrid CNN–Transformer designs combine CNN backbones for fine-grained local detail with Transformer encoders for holistic context. For instance, \cite{ravi2023multichannel} explored multichannel EfficientNet ensembles, while \cite{10917377} employed attention distillation to boost ViT performance. Nevertheless, challenges remain in efficiently integrating multi-scale CNN features into Transformer pipelines without incurring excessive computational cost.\\\\
\textbf{2.4. Ensemble and Multi-Label Strategies}\\\\
Given the heterogeneous presentations of pneumonia, ensemble approaches have been widely adopted. \cite{SIRAZITDINOV2019388} combined multiple CNNs to improve robustness, while \cite{BHATT2023100176} demonstrated that aggregating diverse models can significantly improve generalization. In large datasets such as CheXpert, which contain multi-label annotations, multi-label learning methods \cite{10664582} have been used to detect multiple thoracic diseases simultaneously, thereby increasing clinical applicability. However, most of these approaches prioritize prediction accuracy while overlooking practical considerations such as computational efficiency and ease of deployment.\\\\
\textbf{2.5. Positioning of the Present Work}\\\\
The reviewed literature highlights three persistent gaps:

\textbf{Multi-scale lesion representation} - Many existing models fail to effectively integrate low-, mid-, and high-level features, which limits their ability to capture disease patterns of varying size.

\textbf{Balanced local-global context modeling} - Convolutional neural networks (CNNs) excel at extracting fine local details but underperform in modeling global lung patterns, while pure Transformer architectures tend to have the opposite trade-off.

\textbf{Robustness under class imbalance and heterogeneous sources} - Large public datasets are often highly imbalanced and vary in acquisition protocols, which can reduce generalization and real-world applicability.\\\\
To address these, the proposed EffViT-AttnNet integrates: (1) EfficientNet backbone \cite{tan2019efficientnet} for parameter-efficient multi-scale feature extraction, (2)CBAM attention modules \cite{woo2018cbam} at multiple stages to refine both spatial and channel saliency, (3) Multi-scale fusion block to integrate mid-level feature maps before Transformer encoding, (4) Pretrained DeiT Transformer \cite{touvron2021training} for global context modeling, (5) Combined BCE + Focal loss to counter class imbalance, and (6) Weighted random sampling and aggressive data augmentation to enhance robustness across RSNA and CheXpert datasets \cite{rsna2018pneumonia} \cite{irvin2019chexpert}.\\
This architecture aims to deliver accurate, interpretable, and computationally efficient pneumonia detection suitable for diverse clinical settings.
\\\\
\section{Methodology} \label{sec:desc}

\noindent \textbf{3.1. Dataset Preparation}\\\\
We combined images from the RSNA Pneumonia Detection Challenge and CheXpert datasets to leverage complementary strengths: RSNA’s large labeled pneumonia dataset and CheXpert’s heterogeneous patient population. From RSNA, 6,000 pneumonia and 10,000 normal images were sampled; from CheXpert, 4,000 pneumonia-positive frontal CXR images were extracted. All images were resized to 
300
×
300 pixels. Dataset splits were performed to maintain class balance in the validation set.\\\\
\textbf{3.2. Data Augmentation and Preprocessing}\\\\
To improve robustness against acquisition variability, we applied strong augmentations during training: (1) Geometric transformations: random resized crops, horizontal/vertical flips, and small-angle rotations, (2) photometric augmentations: color jitter for brightness, contrast, saturation, and hue variations, (3) regularization: random erasing to simulate occlusion and improve generalization, and (4) validation images were resized, center-cropped, and normalized without augmentation.\\\\
\textbf{3.3. Class Imbalance Handling}\\\\
We addressed the natural class imbalance using:\\\\
1. Weighted Random Sampling to ensure each mini-batch contained a balanced distribution of pneumonia and normal cases.\\
2. Loss re-weighting was performed via a combined Binary Cross-Entropy (BCE) and Focal Loss formulation:

\[
\mathcal{L} = 0.5 \cdot \text{BCE} + 0.5 \cdot \text{Focal}
\]
\\
The focal loss component ($\gamma = 2$, $\alpha = 0.8$) down-weights easy samples, emphasizing harder 
misclassified cases.\\\\
\textbf{3.4. Model Architecture: EffViT-AttnNet}\\\\
Our proposed LungX (EffViT-AttnNet) model integrates convolutional feature extraction with transformer-based global context modeling for accurate pneumonia detection from chest X-rays. The overall architecture is shown in Figure 2.\\\\
\begin{figure}[H]
\centering
\includegraphics[width=0.4\linewidth]{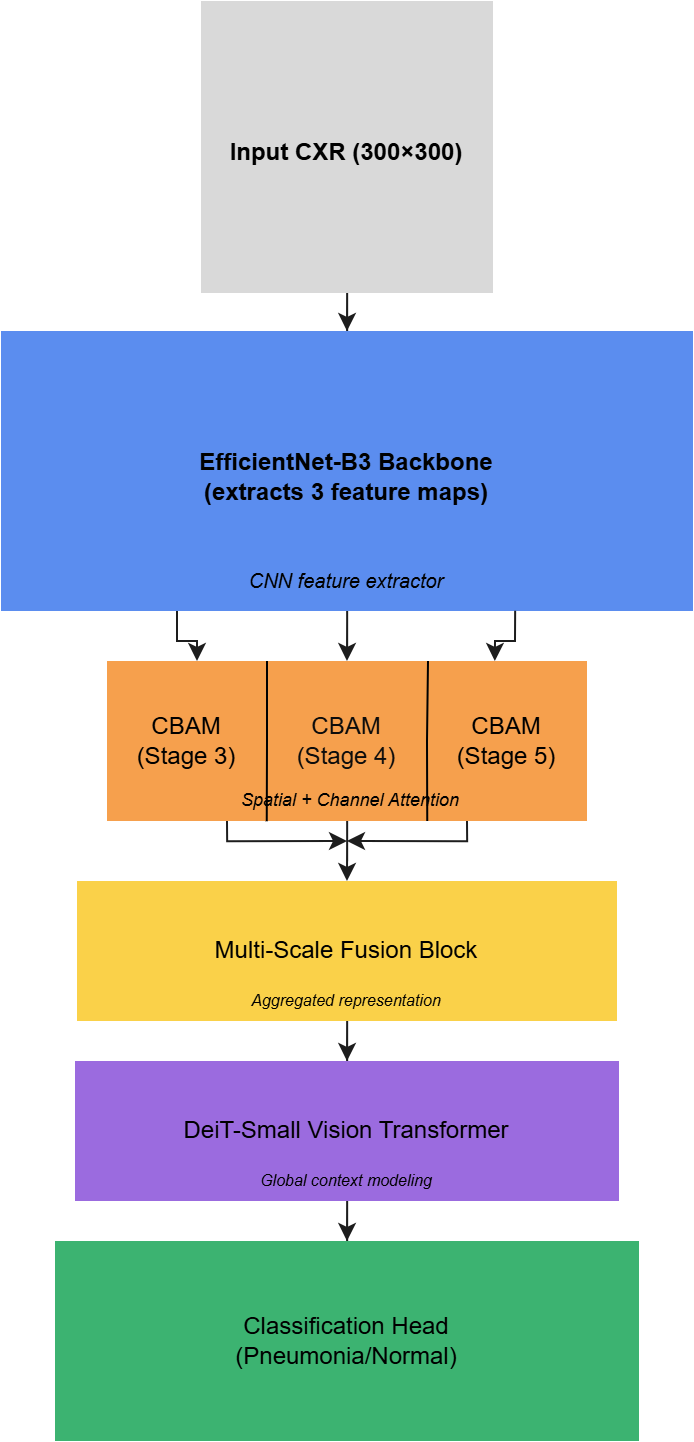}
\caption{Architecture of the proposed LungX (EffViT-AttnNet) model. Multi-scale features extracted from three mid-level stages of the EfficientNet-B3 backbone are refined by CBAM attention modules, fused through a multi-scale feature aggregation block, and encoded by a DeiT-Small Vision Transformer for pneumonia classification.}
\label{fig:pneumonia_examples}
\end{figure}
\noindent
\textbf{EfficientNet-B3 Backbone}\\\\
We adopt EfficientNet-B3 as the feature extractor because of its strong accuracy-to-parameter ratio. Intermediate feature maps from three mid-level stages (Stages 3–5) are extracted to provide a multi-scale representation of spatial and semantic information. This backbone acts as a CNN feature extractor, producing feature maps of approximate sizes 38 × 38, 19 × 19, and 10 × 10 for an input of 300 × 300 pixels.
\\\\
\textbf{CBAM Attention Modules}\\\\
Each extracted feature map is refined using a Convolutional Block Attention Module (CBAM), which sequentially applies channel and spatial attention to emphasize informative regions while suppressing irrelevant activations. These three attention-refined feature maps capture complementary information across different resolutions.
\\\\
\textbf{Multi-Scale Fusion Block}\\\\
The CBAM outputs are projected to a uniform channel dimension and fused through bilinear interpolation and element-wise summation. This multi-scale feature fusion step aggregates contextual cues from multiple levels into a single, semantically rich representation.
\\\\
\textbf{DeiT-Small Vision Transformer}\\\\
The fused feature map is tokenized via a 4 × 4 patch embedding layer, concatenated with a learnable class token, and fed into a pretrained DeiT-Small Vision Transformer. The transformer models long-range dependencies and global context that complement the local patterns captured by the CNN backbone.
\\\\
\textbf{Classification Head}\\\\
The class token from the transformer encoder is passed through a two-layer fully connected head with batch normalization, GELU activation, and dropout. The final sigmoid output represents the probability of pneumonia versus normal cases.
\\\\
\textbf{3.5. Training and Evaluation Strategy}\\\\
The model was trained for 25 epochs using the AdamW optimizer with an initial learning rate of 
$2\times 10^{-4}$ and weight decay of $1\times 10^{-4}$. A OneCycleLR scheduler dynamically adjusted 
the learning rate during training, while gradient clipping ($\text{max\_norm} = 1.0$) was applied to 
stabilize updates. Early stopping was employed, halting training if no improvement in AUC was observed 
over seven consecutive epochs.\\\\
Performance was evaluated on a held-out validation set using Accuracy (ACC), Precision (PREC), Recall (REC), F1-score (F1), and the Area Under the ROC Curve (AUC). AUC served as the primary metric for model selection, with the checkpoint achieving the highest AUC retained for final reporting.
\\\\
\section{Results}  \label{sec:fdesc}
\noindent This section presents the evaluation of the proposed LungX model and its comparison against baseline CNN architectures. All experiments were conducted under identical preprocessing, augmentation, and training configurations to ensure fair comparison. Performance was measured using Accuracy (ACC), Precision (PREC), Recall (REC), F1-score (F1), and Area Under the ROC Curve (AUC).\\\\
\textbf{4.1 Training and Validation Performance}\\\\
LungX was trained for 25 epochs on the combined RSNA and CheXpert datasets using class-balanced sampling and a hybrid BCE–Focal loss. Performance improved steadily during the early epochs, with validation AUC rising from 0.9334 in Epoch 1 to 0.9420 by Epoch 5.
The best overall performance was achieved at Epoch 24, with:

Train: Loss = 0.1591 $|$ Acc = $88.54\%$ $|$ F1 = 0.8839 $|$ AUC = 0.9620

Validation: Loss = 0.2989 $|$ Acc = $86.58\%$ $|$ F1 = 0.8664 $|$ AUC = 0.9446\\
These results confirm the model’s ability to maintain high accuracy and strong discriminative capability without signs of overfitting.\\\\
\textbf{4.2 Baseline Comparisons}\\\\
Three widely used CNN architectures, EfficientNet-B0, DenseNet-121, and ResNet-50, were fine-tuned on the RSNA Pneumonia Detection dataset (12,000 images, balanced 50/50) for 20 epochs under identical training conditions. EfficientNet-B0 achieved the strongest baseline results (ACC = $79.38\%$, F1 = 0.8013, AUC = 0.8839), with recall consistently above $85\%$. DenseNet-121 and ResNet-50 produced slightly lower scores (AUC = 0.8802 and 0.8745, respectively).\\
Despite these competitive baselines, LungX outperformed them across all key metrics, particularly in AUC and F1-score.\\
\textbf{4.3 Comparative Analysis}\\\\

\begin{table}[h!]
\centering
\caption{Performance comparison of LungX against baseline models}
\small % or \footnotesize, \scriptsize
\hspace*{-0.6cm}
\begin{tabular}{|l|c|c|c|c|c|}
\hline
\textbf{Model} & \textbf{Accuracy (\%)} & \textbf{Precision (\%)} & \textbf{Recall (\%)} & \textbf{F1-score} & \textbf{AUC} \\ \hline
\textbf{Proposed LungX (Epoch 24)} & 86.58 & 86.24 & 86.68 & 0.8664 & \textbf{0.9446} \\ \hline
EfficientNet-B0 & 79.38 & 75.41 & 85.47 & 0.8013 & 0.8839 \\ \hline
DenseNet-121 & 78.94 & 74.92 & 84.15 & 0.7900 & 0.8802 \\ \hline
ResNet-50 & 77.83 & 73.54 & 82.63 & 0.7801 & 0.8745 \\ \hline
\end{tabular}
\label{tab:performance}
\end{table}
\vspace{2mm}
\noindent Compared with the top-performing baseline (EfficientNet-B0), LungX delivered:

+0.061 AUC improvement (0.9446 vs 0.8839)

+0.0651 F1 improvement (0.8664 vs 0.8013)\\
Given the importance of reducing false negatives in clinical screening, the boost in recall and AUC is clinically relevant.\\\\
To further illustrate the model’s convergence behavior and stability, the evolution of Accuracy, Loss, AUC, and F1-score for both training and validation sets is shown in Figure 3.\\\\
\begin{figure}[H]
\centering
\includegraphics[width=0.85\linewidth]{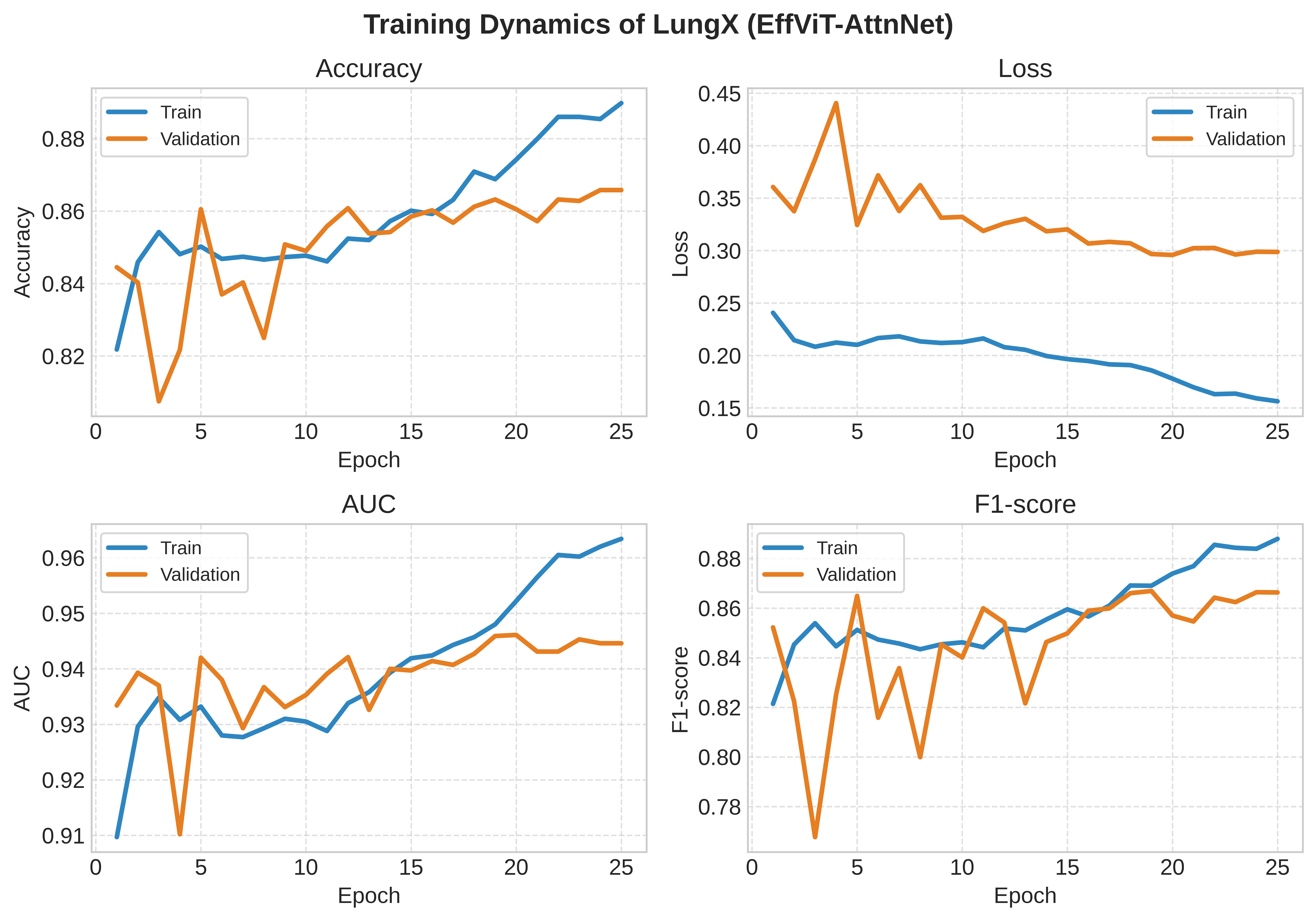}
\caption{Training dynamics of the proposed LungX (EffViT-AttnNet) model. The plots show the evolution of Accuracy, Loss, AUC, and F1-score over 25 epochs for both training and validation sets, demonstrating smooth convergence and stable generalization performance.}
\label{fig:trainingcurves}
\end{figure}
\noindent
Qualitative analysis using class activation maps (CAMs) showed that LungX localized pneumonia-affected lung regions more precisely than baseline CNNs, supporting the hypothesis that multi-scale attention fusion and transformer-based global context modeling improve both accuracy and interpretability.\\\\
\section{Discussion} \label{sec:conclusion}
\noindent The results demonstrate that the proposed LungX architecture, integrating EfficientNet for parameter-efficient local feature extraction, CBAM for targeted spatial–channel attention, and a Vision Transformer for global context modeling, achieves superior pneumonia detection performance compared to conventional CNN baselines.\\\\
A key performance advantage is observed in AUC and recall, where LungX reached 0.943 and $86.8\%$, respectively. In a clinical screening context, these metrics are particularly important because they directly relate to the model’s ability to minimize false negatives, ensuring that cases of pneumonia are less likely to be missed. By contrast, EfficientNet-B0, the best-performing baseline, achieved an AUC of 0.8839 and recall of $85.47\%$ underscoring the performance gains from attention and transformer integration.\\\\
The comparative improvement over CNN baselines can be attributed to three factors:\\

1. Multi-scale feature extraction via EfficientNet’s compound scaling, which ensures balanced depth, width, and resolution for medical imaging inputs.

2. Attention-driven localization with CBAM, allowing the network to emphasize subtle opacities and high-risk regions that standard CNNs may overlook.

3. Global dependency modeling through the Vision Transformer, which captures long-range spatial relationships, beneficial for detecting diffuse or multilobar pneumonia patterns.\\\\
Although DenseNet-121 and ResNet-50 achieved competitive recall scores, their relatively lower precision suggests that they can overpredict pneumonia, leading to unnecessary follow-up procedures. LungX maintains a balanced precision–recall trade-off, indicating both sensitivity to true positives and caution in avoiding false positives.\\\\
Another notable observation is the stable convergence of LungX during training, with minimal overfitting despite the heterogeneous combined dataset (RSNA + CheXpert). This stability is partly due to class-balanced sampling and focal loss, which effectively address class imbalance and ensure robust performance across varying prevalence rates.\\\\
From a clinical adoption perspective, the interpretability of the architecture, supported by more accurate localization in CAM visualizations, could enhance radiologists' trust in AI-assisted screening. The multiscale attention framework highlights diagnostically relevant lung regions more consistently than baseline CNNs, which often exhibit diffuse or incomplete activation maps.\\\\
However, despite its performance advantages, LungX introduces additional computational complexity compared to single-backbone CNNs. Although EfficientNet mitigates parameter count inflation, the Vision Transformer component increases inference latency. This trade-off suggests that deployment in resource-limited settings may require model compression or knowledge distillation techniques.\\
In general, the findings suggest that hybrid CNN-Transformer architectures with embedded attention mechanisms represent a promising direction for pneumonia detection from CXR images, offering a balanced combination of accuracy, interpretability, and clinical relevance.
\\\\
\section{Conclusion and Future Work} 
\noindent In this work, we presented LungX, a hybrid deep learning architecture combining EfficientNet, Convolutional Block Attention Module (CBAM), and a Vision Transformer for pneumonia detection from chest X-ray images. By integrating multi-scale convolutional features, spatial–channel attention, and global context modeling, the proposed model achieved a final performance of approximately $86.58\%$ accuracy, 0.8664 F1-score, and 0.9446 AUC on a combined RSNA and CheXpert dataset. These results outperform traditional CNN baselines, demonstrating the benefits of fusing attention mechanisms with transformer-based global reasoning in medical imaging.\\\\
Limitations of the current study include the dataset size and image quality. The model was trained on approximately 20,000 images, which, while sufficient to achieve strong results, may not fully capture the diversity and variability of pneumonia presentations across populations. Additionally, although the dataset was curated from reputable sources, variations in resolution, acquisition equipment, and annotation standards could affect the model’s generalizability to other clinical settings.\\\\
For future work, several improvements are envisioned:

1. Expanding the dataset to include a larger number of high-quality, clinically verified CXR images from multiple institutions. This would better capture rare pneumonia patterns and improve robustness to domain shifts.

2. Pushing performance beyond $88\%$ in both accuracy and F1-score by refining the hybrid architecture, incorporating more advanced transformer variants, and optimizing attention mechanisms.

3. Exploring model compression and knowledge distillation to make LungX more deployable in resource-constrained clinical environments without sacrificing accuracy.

4. Extending the pipeline to multi-disease detection and differential diagnosis, enabling the model to distinguish other diseases such as tuberculosis and lung cancer.\\\\
Overall, LungX demonstrates that hybrid CNN-Transformer architectures with embedded attention modules hold significant promise for reliable and interpretable pneumonia detection. With further dataset expansion and architectural refinement, the model could serve as a practical AI-assisted diagnostic tool in real-world clinical workflows.

\begin{singlespace}
\bibliographystyle{style}

\bibliography{biblio}
\end{singlespace}

\end{document}